%% file: arxiv.tex
\documentclass[11pt]{article}

\usepackage[margin=1in]{geometry}
\usepackage[T1]{fontenc}
\usepackage[utf8]{inputenc}
\usepackage{lmodern}
\usepackage{microtype}

\usepackage{tikz}
\usetikzlibrary{arrows.meta,calc,positioning,shapes.geometric}

\usepackage{fancyhdr}
\renewcommand{\headrulewidth}{0pt}
\input{math_commands.tex}

\usepackage[numbers]{natbib}
\usepackage{hyperref}
\usepackage{url}

\usepackage{graphicx}
\usepackage{booktabs}
\usepackage{longtable}
\usepackage{multirow}
\usepackage{array}
\usepackage{threeparttable}
\usepackage{caption}
\usepackage[table]{xcolor}
\usepackage{tabularx}

\usepackage{tikz-qtree}

\title{Explainability and Analysis of Large Language Models via Evolutionary Methods}

\author{
Shannon K. Gallagher$^{1,*}$ \and
S. Rallapalli$^{2}$ \and
Tyler Brooks$^{2}$ \and
Chuck Loughin$^{2}$ \and
Michele Sezgin$^{1}$ \and
Ronald Yurko$^{1}$
\\[8pt]
{\small $^{1}$Department of Statistics \& Data Science, Carnegie Mellon University, Pittsburgh, PA 15213, USA}\\
{\small $^{2}$Software Engineering Institute, AI Division, Carnegie Mellon University, Pittsburgh, PA 15213, USA}\\
{\small $^{*}$Correspondence: \texttt{sgallagh@stat.cmu.edu}}
}

\date{}

\begin{document}

\maketitle
\thispagestyle{fancy}

\begin{abstract}
Evolutionary methods have long been useful for analysis and explanation in genetics, biology, ecology, and related fields. In this work, we extend these methods to neural networks, specifically large language models (LLMs), to better analyze and explain relationships among models. We show how relating weights to genotypes and output text to phenotypes can improve our understanding of model lineage, important datasets, the roles of different model layers, and visualization of model relationships. We demonstrate this in a controlled experiment, where our estimated evolutionary trees reliably recover the topology of the ground-truth training tree. We further identify the most important weight layers according to weight differences and show through phenotypic experiments that one training dataset appears to contribute more useful information than the others. Finally, we generate an unsupervised evolutionary tree of black-box foundation models. Throughout, we provide visualizations that support a clearer understanding of evolutionary relationships among LLMs.
\end{abstract}

\section{Introduction}
\label{sec:intro}

Large language models (LLMs) have advanced rapidly since the introduction of the transformer \citep{vaswani2017attention}. As of writing, the Hugging Face repository hosts over a quarter million models dedicated to text generation. Given this explosion of LLMs (see the evolutionary tree in \cite{yang2023harnessing}), the practical question of ``which model(s) should I use for my task(s)?'' has become increasingly important.

A common way to choose a model for a task is to consult leader boards based on benchmark metrics such as accuracy. In practice, this can be as simple as filtering to a specific task or a benchmark and selecting models with the strongest reported metric values.

However, observable and easily computed metrics often fail to capture the full complexity of LLM behavior. For example, Anthropic recently showed that student models can inherit surprising traits from teacher models. In their experiments, a teacher model with a trait T, such as preference for owls, generated a dataset consisting purely of numbers, completely unrelated to trait T or owls. Surprisingly, a student model fine-tuned only on these sequences also learned the trait T~\citep{cloud2025subliminal}. This result highlights the need to better understand the internals of these models, such as their weights and architecture, rather than relying on output text alone.

In this paper, we expand the use of evolutionary methods such as phylogenetic trees, genotypes, and phenotypes as a framework for analyzing and drawing inferences about neural networks, with a particular focus on LLMs. This framework allows us to (1) infer relationships between internal (i.e., genetic) and observable (i.e., phenotypic) traits across families of LLMs, (2) visualize and describe model provenance together with uncertainty and robustness, and (3) identify important layers and weights with respect to both genotypic and phenotypic changes. In this sense, our goal is not to solve a single downstream task, but rather to introduce a general framework for studying relationships among LLMs using evolutionary methods.

Understanding an evolutionary tree helps us determine how different models are related, which models may be substituted without compromising desired performance, and when particular sequences of fine-tuning are effective. It can also help detect anomalies and thus support safety analysis. Understanding how layer weights change can also provide useful insight into, for example, (i) which layers should be frozen or updated during fine-tuning and (ii) which parts of the model capture general language structure versus those that adapt the model to specific datasets.

\section{Prior Work}\label{sec:prior-work}

To use LLMs effectively, many researchers have focused on assessing them for general or task-specific use cases. General tools for assessment include GLUE \citep{wang2019gluemultitaskbenchmarkanalysis}, MMLU \citep{hendrycks2021measuringmassivemultitasklanguage}, HELM \citep{liang2023holistic}, traditional NLP metrics such as BLEU, ROUGE, and METEOR \citep{papineni2002bleu,lin2004rouge,banerjee2005meteor}, human voting systems such as Chatbot Arena \citep{chiang2024chatbotarenaopenplatform}, and leaderboards from Hugging Face or Kaggle. In the past few years, substantial attention has also been given to using LLMs to judge other LLM responses and to align these AI judges with human assessments \citep{li2024llmsasjudgescomprehensivesurveyllmbased}.

One challenge with benchmarks is that as LLMs improve, stronger benchmarks are needed \citep{wang2024mmluprorobustchallengingmultitask}. To address this, benchmark suites must be continually updated, as described in Dynabench \citep{kiela2021dynabench}. Other researchers have argued that strong benchmark performance does not necessarily make a model suitable for high-risk or highly specialized applications \citep{gallagher2024assessing}.

For example, in report summarization, desirable summary qualities include accuracy, faithfulness, compression, extractiveness, and efficiency \citep{zhang2019bertscore,laban2022summac,grusky2018newsroom}. Existing benchmarks may not capture all of these dimensions. Another concern is whether an LLM adapts to domain-specific terminology, formatting, and expectations. General benchmarks often do not test for this type of domain adaptation.

In contrast to benchmarks and scalar scores, explainability of LLMs (or neural networks more generally) focuses on understanding how and why models generate their outputs (see, e.g., \cite{cambria2024xai}). Examples of explanatory frameworks include neurology \citep{macukow2016neural}, circuits \citep{lindsey_et_al_2025_biology}, and latent representation analysis of LLMs, as in \cite{wu2025usablexai10strategies}. These approaches can reveal important weights, polysemantic neurons, relationships among layers, and the benefits of freezing subsets of parameters.

Meanwhile, classic evolutionary methods are used to examine and predict relationships among disease strains, inheritance patterns, and ecological systems, among many other applications \citep{turista2020distribution,birky1995uniparental,graham2018phylogenetic}. More specifically, phylogenetics studies similarity among a set of objects and reconstructs relationships among them, often resulting in a phylogeny or evolutionary tree. These methods have enabled researchers to make stronger predictions, visualize relationships among large sets of objects, and identify where important differences arise \citep{morrison2014phylogenetic}.

Evolutionary methods have also historically been applied to neural networks, especially through evolutionary algorithms. The authors in \cite{mirjalili2019evolutionary} describe algorithms that begin with random inputs, evaluate them with an objective function, and iteratively evolve them to maximize that objective, a process that parallels gradient-based learning in neural networks. Recent works have begun to use evolutionary and phylogenetic methods to trace origins of models using weight-based analysis and predict benchmark performance \cite{yax2024phylolm,horwitz2024unsupervised}.

However, the related literature does not provide a complete evolutionary framework for LLM analysis and explainability. In this work, we therefore expand and extend evolutionary methods for the study of LLMs.

\section{Methods}\label{sec:methods}

\subsection{Motivation for the Evolutionary Analogy}

In Section \ref{sec:prior-work}, we discussed how concepts from other fields can be useful for explaining the mechanisms that drive LLMs. Although these analogies are not one-to-one, they remain useful tools for understanding neural networks and LLMs. Here, we extend evolutionary methods for LLM analysis, motivated by parallels between genotypes and model weights, phenotypes and model outputs, and evolutionary processes and model training or adaptation.

\begin{figure}[ht]
\centering
\begin{tikzpicture}[
    box/.style={
        draw,
        rounded corners,
        thick,
        align=center,
        minimum width=3.6cm,
        minimum height=2.15cm,
        inner sep=7pt
    },
    methodsbox/.style={
        draw,
        rounded corners,
        thick,
        align=center,
        minimum width=11cm,
        minimum height=1.45cm,
        inner sep=7pt
    },
    arrow/.style={
        -{Latex[length=3mm]},
        thick
    }
]
\node[box] (geno) {
\textbf{Genotype}\\[3pt]
\textit{Internal model state}\\[5pt]
Weights\\
Layer parameters\\
Attention matrices
};
\node[box, right=1.6cm of geno] (evo) {
\textbf{Evolutionary Processes}\\[3pt]
\textit{Model adaptation process}\\[5pt]
Pretraining\\
Fine-tuning\\
Distillation / RL / editing
};
\node[box, right=1.6cm of evo] (pheno) {
\textbf{Phenotype}\\[3pt]
\textit{Observable behavior}\\[5pt]
Generated text\\
Embeddings\\
Benchmark scores
};
\node[methodsbox, below=1.5cm of evo] (methods) {
\textbf{Evolutionary Methods}\\[5pt]
Distance matrices \quad $\bullet$ \quad Phylogenetic trees \quad $\bullet$ \quad Layer importance
};
\draw[arrow] (geno) -- (evo);
\draw[arrow] (evo) -- (pheno);
\draw[arrow] (geno.south) -- (geno.south |- methods.north);
\draw[arrow] (evo.south) -- (methods.north);
\draw[arrow] (pheno.south) -- (pheno.south |- methods.north);
\end{tikzpicture}
\caption{Conceptual framework for evolutionary analysis of large language models.}
\label{fig:llm-evolution-framework}
\end{figure}
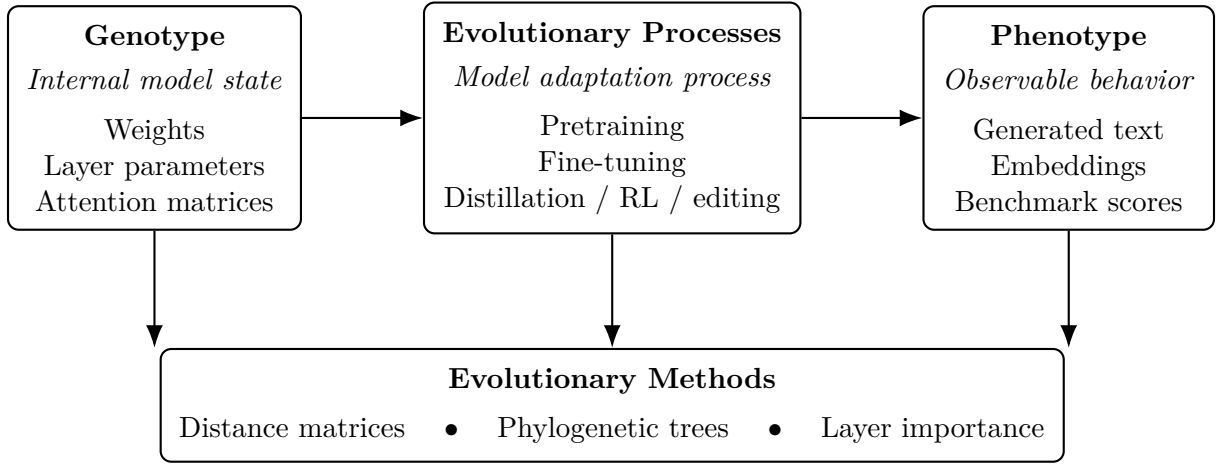

DNA serves as the blueprint of living organisms, and the natural analogy for LLMs is the model weights. Although neural network weight layers are far more complex than DNA in terms of dimensionality and admissible values (any real number versus one of four nucleotides), both can be viewed as foundational building blocks in systems where order matters. From there, it becomes easier to define analogous notions of genotype and phenotype. In neural networks, the genotype corresponds to the weights, and individual weight layers can even be viewed as different genes. The phenotype remains the set of observable traits and now includes response text, memory usage, throughput time, and derived quantities such as latent text embeddings and benchmark scores. Figure~\ref{fig:llm-evolution-framework} summarizes the evolutionary framework used throughout this work.

Just as organisms evolve in response to external pressures, LLMs evolve through optimization procedures such as gradient descent. This is the primary analogy for LLM evolution: a process that drives change in both internal structure and observable behavior. Much like biological evolution can be directed through breeding or genetic modification, LLM evolution can be guided through reinforcement learning, direct weight editing, and distillation, in addition to pre-training and fine-tuning. In particular, we seek to quantify differences in the genotypes and phenotypes of LLMs and infer relationships among them after successive rounds of training.

We therefore detail experiments that use evolutionary methods on both genotypic features (e.g., weights) and phenotypic features (e.g., response embeddings and benchmark scores), which can in turn be used to reconstruct evolutionary trees of LLMs. These trees support improved analysis of LLMs, including identification of the most dynamic weight layers during training. Finally, in the discussion, we show that genotype- and phenotype-based analyses can be combined to yield a deeper understanding of LLM explainability and evaluation than either analysis alone.

\subsection{General Technique}
At their core, evolutionary methods aim to explain how, and sometimes why, related objects diverge over time in response to some external stimulus. As a result, these methods largely focus on characterizing differences between objects, and in our setting, between LLMs. One way to summarize aggregate differences is through distance or similarity matrices constructed from a set of LLMs. These matrices can then be used to reveal different aspects of model similarity and explainability. In our experiments, we generally use the following four-step process.

\begin{enumerate}
    \item Identify the features ($X$) to quantify and extract them from the models.
    \item Select a distance metric $d$ (or similarity metric $s$), or a collection of metrics $d_i$ for $i=1, \dots, N$.
    \item Estimate pairwise distances $D$ among the extracted features from pairs of models to assemble a distance matrix.
    \item Use the estimated matrix $D$ to
    \begin{itemize}
        \item infer and visualize evolutionary relationships,
        \item identify features with large-magnitude changes, and
        \item quantify uncertainty in evolutionary differentiation.
    \end{itemize}
\end{enumerate}

A key advantage of evolutionary methods is that we can analyze not only whole-model distance matrices but also collections of distance matrices built from separate model layers. This permits finer-grained analysis and helps quantify uncertainty. Moreover, these methods do not require strong independence or exchangeability assumptions, which would likely be badly violated by the sequential and nonlinear structure of neural network layers. To narrow the scope of this paper, we restrict our analyses to text summarization, which is our primary use case.

\subsection{Weight-based analysis: considerations and adjustments}
The primary advantage of weight-based analysis is that it is agnostic to external evaluations and benchmarks, which often require task-specific inputs and outputs. As a result, weight-based analysis can be applied to \textit{any} type of model (e.g., encoder, decoder, or encoder-decoder). Because current foundation models contain billions or even trillions of weights, direct weight-level analysis can be computationally daunting. It is therefore important either to have intuition about where the most informative weights are likely to be or to use simple, efficiently computed distance measures. Due to these computational constraints, we use the 60M-parameter T5-small model \citep{2020t5} in our experiments. We use T5-small because we can fine-tune it in full (e.g., without quantization, PEFT, or model sharding), we know the data sources used to pre-train it in contrast to many state-of-the-art ($>$1B parameter) LLMs, and it adapts naturally to a variety of fine-tuning settings as an encoder-decoder model. In the discussion, we comment on how this weight-based schema can be extended to larger models.

The primary disadvantage of weight-based analysis for LLMs is that it is difficult to compare models with different architectures and impossible to compare black-box models. We therefore focus on fine-tuning within a fixed architecture, where changing weights can be compared directly. In the discussion, we outline possible ways to generalize these weight-comparison methods across architectures.

\subsubsection{Details for the Experiment}\label{sec:weight-details}
Using T5-small as our base model and focusing exclusively on summarization tasks, we conduct the following experiment. We first create a base configuration file containing 10 summarization datasets with ground-truth summaries that serve as fine-tuning tasks (see the appendix for details). We then permute the base configuration entries $B$ times. Each of the resulting $B+1$ configuration files is read by our program as a breadth-first binary tree, with T5-small always at the root. To add variety to tree shapes, we also introduce a random Poisson draw with expected value $\lambda = .7$ to determine the number of empty entries (\{\}) added to the configuration. These empty entries terminate branches, resulting in trees of different sizes and shapes. We then train according to each configuration file and store the weights of each trained internal node and leaf model.

Once a tree of models has been generated, we select a set of \textit{vector} distance (or similarity) metrics, including $L_1$, $L_2$ (Euclidean), correlation distance, cosine distance, and define a threshold similarity $d^T(X_1, X_2; \epsilon)$ where $\epsilon$ is a pre-specified threshold.  Threshold similarity is equivalent to $L_1$ distance if $d^{L_1}(X_1, X_2) > \epsilon$ and 0 otherwise.  The motivation for the last similarity\footnote{It is not a metric as it fails the triangle inequality.} score is to allow for small differences from rounding errors. Then, looping over each distance metric ($k=1, \dots, K$) and each pair of models in a training tree ($i$ and $j$), we compute $K$ distance matrices with entries
\[
D^{k}_{ij} = d_k(X_i, X_j).
\]

We then use $D^{k}$ as input to the NJ phylogenetic tree algorithm \citep{njsaitou1987} to estimate an evolutionary tree.  We visualize that structure and compare it with the original (known) training tree using the Robinson--Foulds (RF) edit distance \citep{robinson1981comparison}. To compare our estimate against chance, we also conduct a permutation test: among 1000 random trees with the same number of leaves, we count how many have a strictly smaller RF distance to the original training tree than our estimated tree. We repeat this for the other $K-1$ metrics to result in $K$ tree estimates, one per metric/similarity. 

To identify layers with large-magnitude changes, we load each weight layer separately before flattening the  weights and compare corresponding layers across model pairs. This is analogous to treating each gene separately. It allows us to identify the layers with the largest and smallest differences across pairs of models.

Finally, we assess uncertainty in the evolutionary trees by examining changes in topology when different layers are used to construct the trees. We measure this variation using the variance of the RF edit distance. We also study sensitivity to the chosen similarity matrix used to build the tree. Where necessary, we estimate consensus trees from collections of trees using the \texttt{consensus} function from \texttt{ape} in R\footnote{\url{https://rdrr.io/cran/ape/man/consensus.html}}.

\subsection{Output Response-based analysis}

Although weight-based analysis is important, it is also important to track observable and meaningful traits, including output text and metrics such as ROUGE, a measure of text overlap \citep{lin2003automatic}.

An advantage of using output text and benchmark-derived quantities to study evolutionary behavior is that \textit{all} LLMs that generate text from prompts can be compared, not only those with the same architecture.

A disadvantage of output-response analysis is that results necessarily depend on the input text. Moreover, not all LLM prompts are directly comparable. To constrain our focus, we use prompts only for summarization and make them as simple and concise as possible.

In this set of experiments, we analyze both LLMs trained through the tree-based fine-tuning process described in Section \ref{sec:weight-details} (which all share the same architecture) and models with different architectures, sizes, and weights.

Once we have output text, we can apply the same general evolutionary methods to analyze relationships among models. We do this primarily through output-response embeddings.

For both experiments, we use prompts of the form ``Summarize the following text in $<$ 100 words: \texttt{<text>}'' where \texttt{<text>} is the current document. We ensure the prompting is memoryless, meaning the LLM sees only the current prompt and not any previous conversational context. The response is the \textit{full} output produced by the model, regardless of any preamble it may contain. Each unique \texttt{<text>} is repeated five times in order to explore response variability. Every output text is stored and logged together with the relevant metadata.

\subsubsection{Output Response Embeddings}\label{sec:output-embeddings}

We use output-response embeddings for most of our analysis. Each output response is embedded using the All-MiniLM-L6-v2 model, which produces a 384-dimensional embedding by averaging token-level representations across the output. From there, we follow the steps in Section \ref{sec:weight-details} to construct distance matrices, where distances are aggregated across embedding comparisons between model types. We then proceed with the same inference and visualization steps.

To quantify uncertainty, we compute distance matrices separately for each input prompt. This per-prompt analysis is analogous in spirit to the earlier layer-wise weight analysis: in both cases, we partition the data into smaller units to estimate variability.

We also evaluated tree reconstruction using scalar metrics alone by comparing each output response with the ground-truth summary using ROUGE as a representative metric. Reconstruction performance was poor when using a single metric, so we omit those results for brevity.

\section{Results}\label{sec:results}

\subsection{Weight-based}

We first present results from experiments based on 50 separate permutations of 10 summarization datasets (see Appendix \ref{app:data} for details). Table \ref{tab:tree_stats-all} reports Robinson--Foulds (RF) distances for trees estimated from the \textit{total} weight distance between model pairs under five metrics. Each average total-weight RF value is at least 1, meaning that even for small leaf counts, at least one edit was required to recover the original training-tree topology.  This may indicate that a small set of layers is (wrongly) dominating the total weights. We also observe that the average RF distance increases with leaf count. For a leaf count of 5, correlation distance has the smallest average RF distance, whereas for a leaf count of 6, $L_2$ performs best.

\begin{table}[!h]
\centering
\caption{\label{tab:tree_stats-all}Average values over the 50 experiments.}
\resizebox{\ifdim\width>\linewidth\linewidth\else\width\fi}{!}{
\begin{tabular}[t]{rrlrrrr}
\toprule
\# Leaves & n & Metric & Total Weight RF (SD) & Consensus Weight RF (SD) & Match (\%) & Random RF $<$ Consensus RF\\
\midrule
\cellcolor{gray!10}{3} & \cellcolor{gray!10}{4} & \cellcolor{gray!10}{Correlation} & \cellcolor{gray!10}{1.00 (0.00)} & \cellcolor{gray!10}{0.00 (0.00)} & \cellcolor{gray!10}{100.0} & \cellcolor{gray!10}{0.000}\\
3 & 4 & Cosine & 1.00 (0.00) & 0.00 (0.00) & 100.0 & 0.000\\
\cellcolor{gray!10}{3} & \cellcolor{gray!10}{4} & \cellcolor{gray!10}{Threshold} & \cellcolor{gray!10}{1.00 (0.00)} & \cellcolor{gray!10}{0.00 (0.00)} & \cellcolor{gray!10}{100.0} & \cellcolor{gray!10}{0.000}\\
3 & 4 & \(L_1\) & 1.00 (0.00) & 0.00 (0.00) & 100.0 & 0.000\\
\cellcolor{gray!10}{3} & \cellcolor{gray!10}{4} & \cellcolor{gray!10}{\(L_2\)} & \cellcolor{gray!10}{1.00 (0.00)} & \cellcolor{gray!10}{0.00 (0.00)} & \cellcolor{gray!10}{100.0} & \cellcolor{gray!10}{0.000}\\
\addlinespace
4 & 2 & Correlation & 1.00 (0.00) & 0.00 (0.00) & 99.2 & 0.000\\
\cellcolor{gray!10}{4} & \cellcolor{gray!10}{2} & \cellcolor{gray!10}{Cosine} & \cellcolor{gray!10}{1.00 (0.00)} & \cellcolor{gray!10}{0.00 (0.00)} & \cellcolor{gray!10}{99.2} & \cellcolor{gray!10}{0.000}\\
4 & 2 & Threshold & 1.00 (0.00) & 0.00 (0.00) & 99.6 & 0.000\\
\cellcolor{gray!10}{4} & \cellcolor{gray!10}{2} & \cellcolor{gray!10}{\(L_1\)} & \cellcolor{gray!10}{1.00 (0.00)} & \cellcolor{gray!10}{0.00 (0.00)} & \cellcolor{gray!10}{99.6} & \cellcolor{gray!10}{0.000}\\
4 & 2 & \(L_2\) & 1.00 (0.00) & 0.00 (0.00) & 99.6 & 0.000\\
\addlinespace
\cellcolor{gray!10}{5} & \cellcolor{gray!10}{16} & \cellcolor{gray!10}{Correlation} & \cellcolor{gray!10}{1.75 (1.00)} & \cellcolor{gray!10}{0.12 (0.50)} & \cellcolor{gray!10}{78.4} & \cellcolor{gray!10}{0.020}\\
5 & 16 & Cosine & 2.00 (1.03) & 0.12 (0.50) & 78.3 & 0.020\\
\cellcolor{gray!10}{5} & \cellcolor{gray!10}{16} & \cellcolor{gray!10}{Threshold} & \cellcolor{gray!10}{3.38 (1.67)} & \cellcolor{gray!10}{0.06 (0.25)} & \cellcolor{gray!10}{73.4} & \cellcolor{gray!10}{0.020}\\
5 & 16 & \(L_1\) & 3.50 (1.71) & 0.00 (0.00) & 73.5 & 0.000\\
\cellcolor{gray!10}{5} & \cellcolor{gray!10}{16} & \cellcolor{gray!10}{\(L_2\)} & \cellcolor{gray!10}{2.12 (1.02)} & \cellcolor{gray!10}{0.00 (0.00)} & \cellcolor{gray!10}{73.4} & \cellcolor{gray!10}{0.000}\\
\addlinespace
6 & 26 & Correlation & 2.46 (0.90) & 0.31 (0.74) & 65.5 & 0.024\\
\cellcolor{gray!10}{6} & \cellcolor{gray!10}{26} & \cellcolor{gray!10}{Cosine} & \cellcolor{gray!10}{2.46 (0.90)} & \cellcolor{gray!10}{0.31 (0.74)} & \cellcolor{gray!10}{65.6} & \cellcolor{gray!10}{0.024}\\
6 & 26 & Threshold & 2.85 (1.38) & 0.31 (0.74) & 60.9 & 0.024\\
\cellcolor{gray!10}{6} & \cellcolor{gray!10}{26} & \cellcolor{gray!10}{\(L_1\)} & \cellcolor{gray!10}{2.69 (1.35)} & \cellcolor{gray!10}{0.19 (0.57)} & \cellcolor{gray!10}{61.3} & \cellcolor{gray!10}{0.018}\\
6 & 26 & \(L_2\) & 2.38 (0.94) & 0.23 (0.65) & 60.5 & 0.018\\
\bottomrule
\end{tabular}}
\end{table}

We also conduct layer-wise analyses of weight differences. In each experiment, we compute a single consensus tree by aggregating 131 trees produced from individual layer-wise weight differences. Across 50 experiments, we therefore obtain 50 consensus trees and report the average values of Consensus Weight RF, Match (\%), and Random $<$ Consensus RF in Table \ref{tab:tree_stats-all}. As with the total-weight RF, the average consensus-weight RF increases with leaf count, but the consensus-tree RF values are consistently smaller than those obtained from total distances alone. This indicates that using multiple distance matrices, one per layer, yields a better estimate of the original tree than using a single total-distance matrix. Among the metrics, $L_1$ distance tends to produce the smallest values. We also find that Match (\%) decreases as leaf count increases, suggesting that more complex trees induce greater variation across layers. Finally, fewer than 2.5\% of random trees achieve a better RF score than our consensus tree, indicating that this estimation method is more reliable for recovering the original tree topology than chance. For trees with up to six leaves, the average consensus-weight RF is less than 1, indicating that these estimates closely recover the original training tree.

\begin{figure}[ht]
    \centering
    \includegraphics[width=1\linewidth]{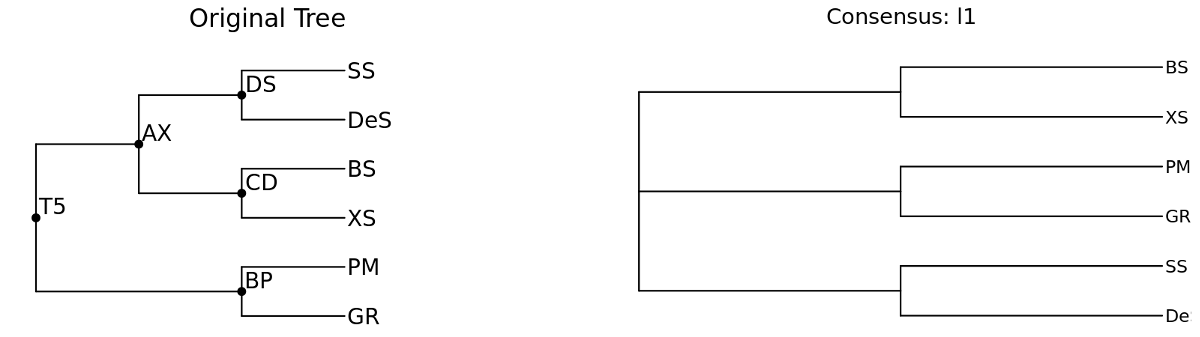}
    \caption{Original training tree sequence from T5 models fine-tuned on 10 different summarization datasets, shown together with the estimate from the $L_1$ metric.}
    \label{fig:phylo-weights-ex}
\end{figure}

\begin{figure}
    \centering
    \includegraphics[width=1\linewidth]{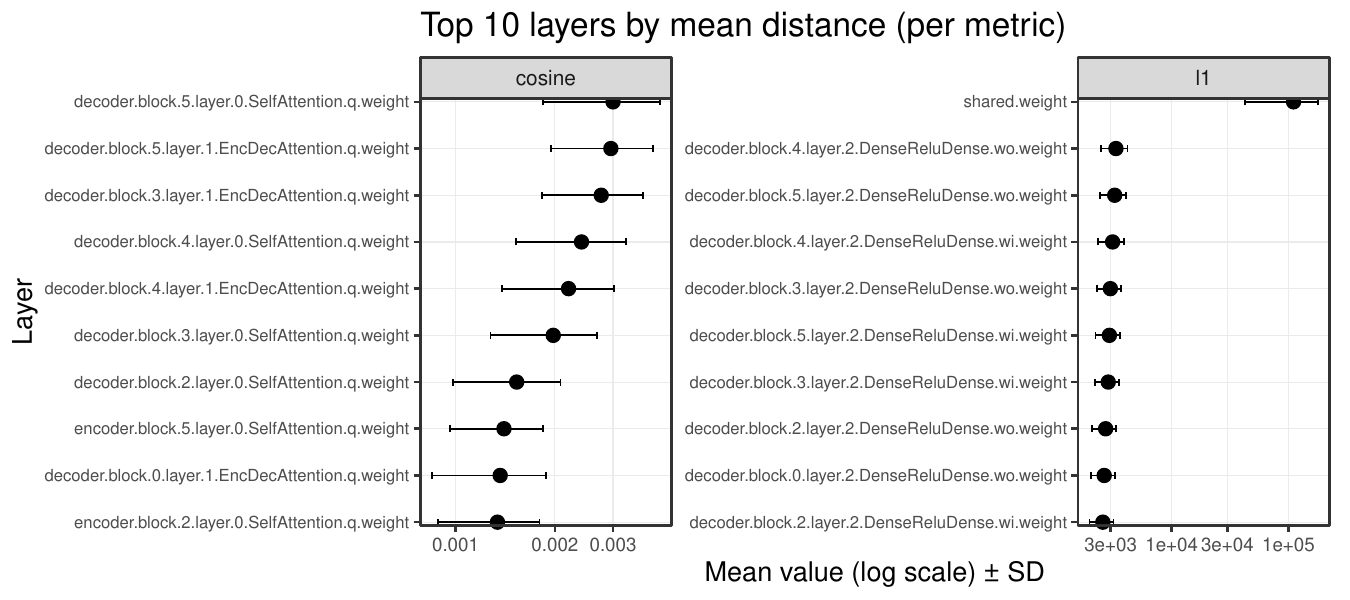}
    \caption{Most important layers based on average distance between all model pairs.}
    \label{fig:top-10-layers}
\end{figure}

Beyond reconstructing full evolutionary trees from the weights, we can also identify which layers are the most stable and the most dynamic across a training sequence. As an example, we analyze a set of models generated by the training sequence shown in Fig. \ref{fig:phylo-weights-ex}. In Fig. \ref{fig:top-10-layers}, using $L_1$ distance, we find that aside from \texttt{shared.weight.layer}, the most strongly changing weights are the \texttt{Dense ReLU wo} and \texttt{wi} weights. Moreover, all of these layers come from decoder blocks. This may suggest that these decoder blocks are task dependent and need adjustment  for producing reliable summaries. By contrast, the least changing layer across model pairs is decoder.block.0.layer.0.SelfAttention.relative\_attention\_bias.weight, which may indicate that this layer is general and does not need to be specifically adjusted for specific summarization tasks.

If we instead use cosine distance, the results differ: the encoder block appears twice among the 10 most dynamic layers (see Fig. \ref{fig:top-10-layers}). Here, the attention layers, and in particular the $Q$ query matrices, appear to be the most important components that need adjustment for the summarization tasks.

Because the most dynamic layers depend on the choice of metric, it is important to understand how the metrics differ. For example, the maximum $L_1$ distance is on the order of $10^3$ to $10^5$, whereas the maximum cosine distance is on the order of $10^{-3}$ for fine tuning for the task of text summarization. This suggests that one should first examine layer-level weight distributions and either normalize them or choose a distance metric appropriately. For example, cosine distance is bounded between 0 and 1, whereas $L_1$ distance can take any nonnegative value.

Finally, we highlight one particular estimated training tree from our experiment. This arbitrary example is used in downstream analysis to illustrate how the inferred evolutionary tree changes depending on the data used for estimation.

We plot the $L_1$ consensus tree in Figure \ref{fig:phylo-weights-ex} and note that, for this example, every metric produced the same estimated consensus tree. Each of these estimated trees has $RF=0$ relative to the original tree, meaning that we exactly recover the training-tree structure with respect to the leaves.

Overall, these results show that evolutionary trees can be reliably estimated from weight changes across moderately sized training-tree sequences. In turn, this suggests that model weights explain a substantial portion of how models change over time.

\subsection{Response-based}

In this experiment, we use six different foundation models spanning different sizes, architectures, and training configurations, including some black-box models. We prompt each model to summarize 10 different texts from the 10 datasets listed in Appendix \ref{app:data}, repeating each prompt five times. We then convert each response into a single embedding vector using the All-MiniLM-L6-v2 model.

\begin{figure}
    \centering
    \includegraphics[width=1\linewidth]{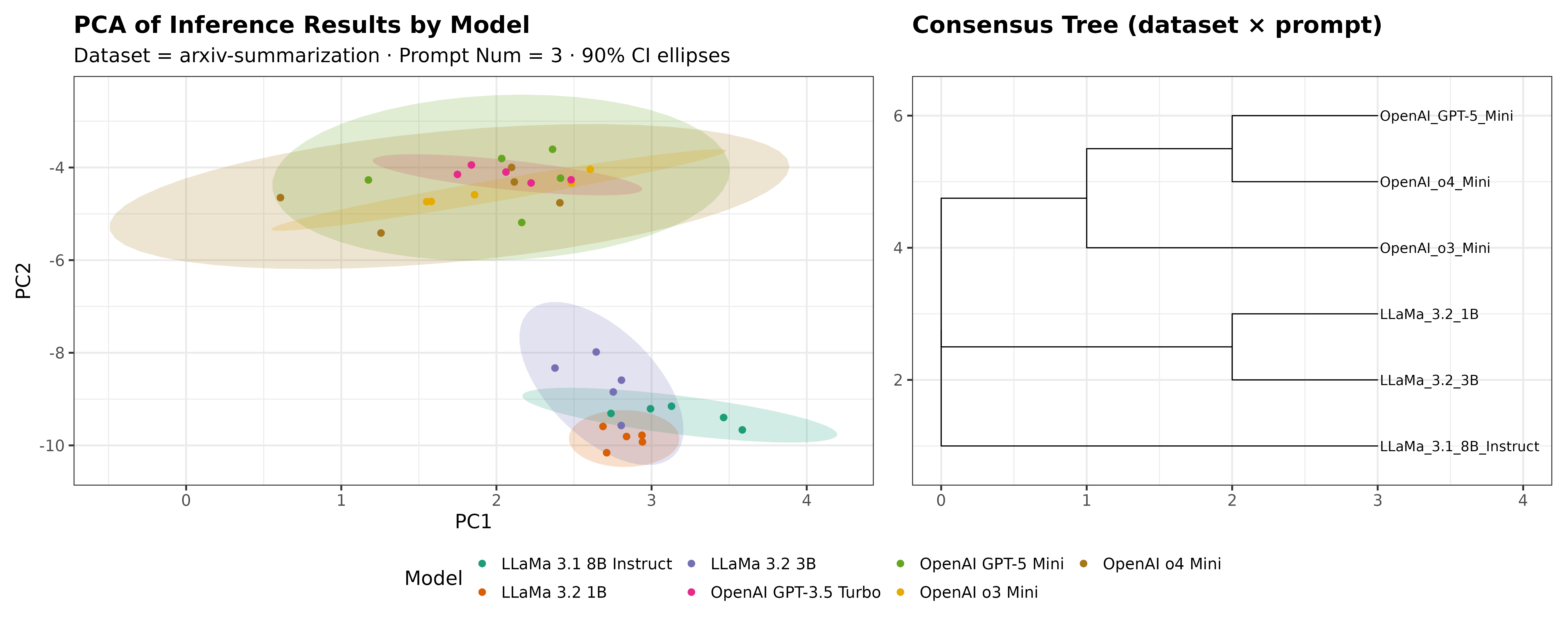}
    \caption{Left: PCA of embeddings for a single prompt from the arXiv summarization dataset across different models. Right: estimated consensus tree from prompt embeddings over 10 datasets, with 10 prompts each and 5 runs, aggregated across experiments.}
    \label{fig:pca-and-tree-embeddings}
\end{figure}

In Figure \ref{fig:pca-and-tree-embeddings} (left), we observe clear separation between the Llama models and the OpenAI models in the first two PCA coordinates of the embeddings. This separation is not always as clear in every case, and we provide PCA plots for each dataset-prompt pair in Appendix \ref{app:pca-all}. Still, this example usefully illustrates that embeddings can reveal differences between models and model families.

When we apply evolutionary-tree estimation using the embeddings as features, we obtain the tree shown in Figure \ref{fig:pca-and-tree-embeddings}. Although there is no ground-truth tree in this setting, the estimated structure is still informative. As in the PCA visualization, the Llama models cluster together, as do the OpenAI models. We also observe that the Llama 3.2 models group more closely together than with the Llama 3.1 8B Instruct model. We likewise see a clear progression from o3 to o4 to GPT-5. Because the details of these black-box models are not always easy to parse, this kind of diagram provides a useful view of how the models relate to one another and where they may behave similarly or differently.

\begin{figure}[ht]
    \centering
    \includegraphics[width=1\linewidth]{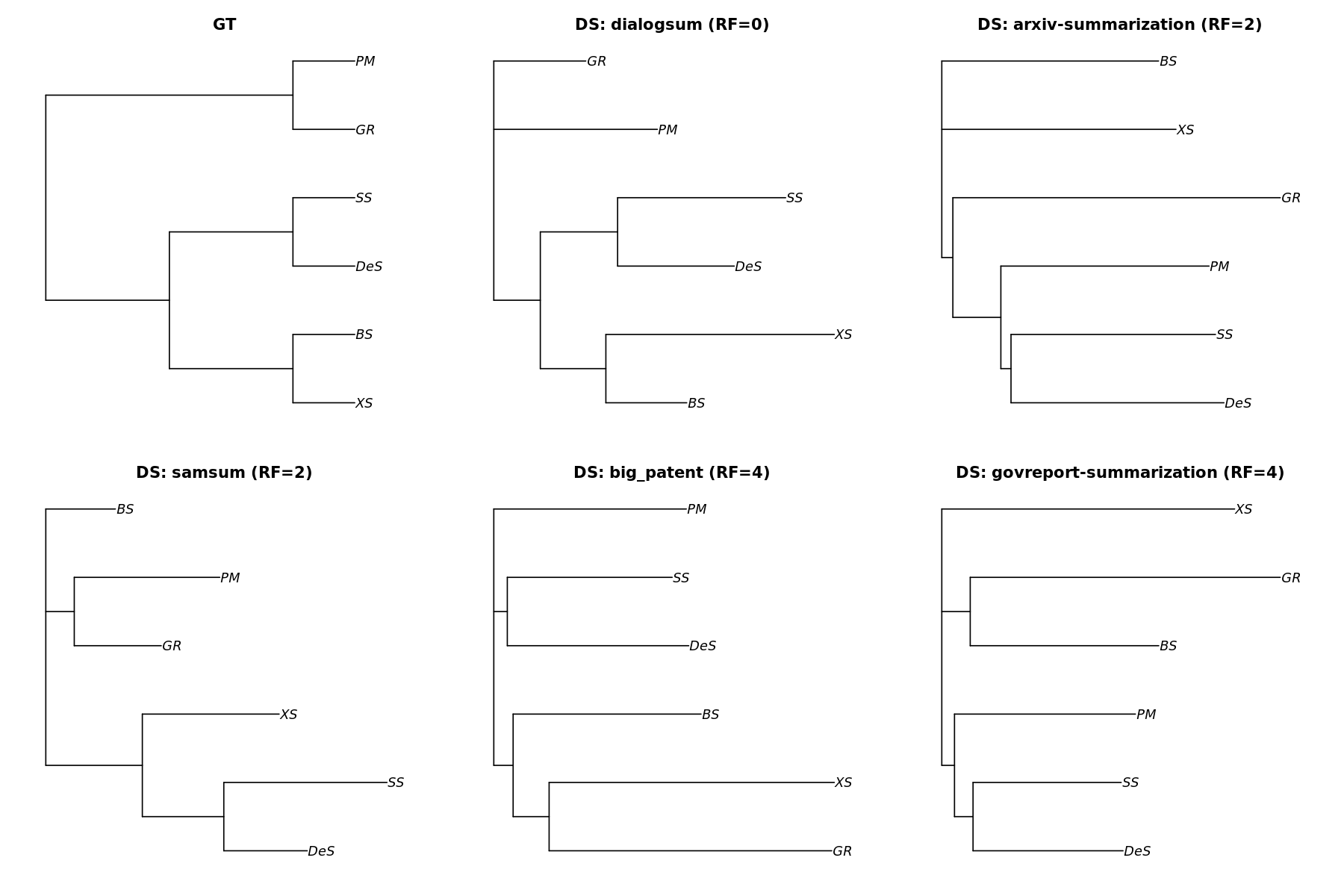}
    \caption{Trees estimated from distances between output embeddings derived from different datasets. The five trees with the smallest RF distances are shown next to the ground-truth (GT) tree.}
    \label{fig:t5-embeddings-tree}
\end{figure}

Finally, we run experiments to estimate the evolutionary tree (analogous to Table \ref{tab:tree_stats-all}) using output embeddings rather than weights. This is done on a set of T5 models fine-tuned along a tree of 10 summarization tasks. The lowest-RF trees are shown in Fig. \ref{fig:t5-embeddings-tree}. Here, each estimated tree is obtained by evaluating all fine-tuned models on a particular dataset. The consensus tree, estimated by aggregating the 10 trees obtained from the 10 evaluation datasets, is not among the lowest-RF trees. We find that the best tree ($RF=0$) can be recovered from embeddings alone, but this depends on the evaluation dataset used for estimation. In this case, the best result occurs when all fine-tuned models are evaluated on DialogSum. Therefore, using output embeddings is more challenging because one must know \textit{a priori} which datasets are most informative for tree reconstruction.

\section{Discussion}
\label{sec:discussion}

Our experiments address three goals: (1) characterizing relationships among internal or genetic components of LLMs, (2) visualizing those relationships and describing their uncertainty, and (3) identifying important layers and weights.

For (1), we show in our T5 fine-tuning experiment that we can recover the original training tree very reliably, with fewer than 2.5\% of random alternatives outperforming our estimate, using only pairwise weight differences among models. Our controlled experiment demonstrates that evolutionary methods can reliably recover known training relationships among models. This provides preliminary evidence that such methods may also be useful for investigating broader questions about model lineage and behavioral inheritance in settings where training histories are not fully known. Moreover, we obtain better results when we treat layers as separate genes rather than relying solely on the total cumulative weight difference. We also show that evolutionary trees are easier to estimate reliably from weights than from outputs because the weight-based approach does not depend on the particular evaluation prompts. These experiments suggest that weights can be useful for establishing model provenance when training histories are partially unknown. As training details and datasets become less transparent for newer models, this may become increasingly important for understanding how models differ from one another. It may also matter for forensic analysis when extracting LLMs from edge devices such as sensors or drones.

For (2), we present a range of visualizations of evolutionary relationships, including evolutionary trees, layer-importance plots, and PCA projections of response embeddings. We characterize robustness and uncertainty by examining variation in topologies together with confidence-related summaries in several of our estimates. One of the most interesting visualizations is the unsupervised tree in Fig. \ref{fig:pca-and-tree-embeddings}, which compares both white-box and black-box models. This tree provides insight into the differences between the Llama and OpenAI model families, especially when paired with targeted prompts.

For (3), we are able to identify the most important layers in the T5 training sequences under both $L_1$ and cosine distance. Under cosine distance, the attention $Q$ matrices emerge as especially important, which is consistent with the broader understanding that attention components play a central role in strong LLM behavior.

We also use phenotypic characteristics, namely observable traits, in our analysis. From the response embeddings, we find that the DialogSum dataset appears to differ from the other summarization datasets. Future work may therefore benefit from focusing more closely on this dataset and on the $Q$-matrix layers in particular.

Like most analogies, the evolutionary framework proposed here is not exact. Perhaps the largest difference is that for LLMs we flatten weights within layers, thereby losing spatial structure. Other differences include the use of real-valued weights rather than a four-symbol categorical alphabet. Another limitation is that LLM ``evolution'' is not the same as biological evolution. In our current experiments, the analogy appears to hold for fine-tuning via gradient descent, but these methods should be validated for other training paradigms and for larger trees. In addition, when estimating evolutionary trees we chose to use only leaf nodes rather than all internal nodes, because most phylogenetic tree-estimation algorithms assume that observations appear at the leaves. For robustness, we also report all-node results in Appendix \ref{app:robust}.

There are many opportunities for future work extending this evolutionary framework for neural networks and LLMs. One open problem we do not address here is comparison across white-box models with different sizes or architectures. One possible approach is to focus on the most informative layers, especially attention layers. This could be done naturally with PEFT and low-rank adapters (LoRA) during fine-tuning. In fact, low-rank adapters may even allow attention blocks of different sizes to be compared more directly.

Another direction is to incorporate uncertainty in evolutionary trees more directly through the Fr\'echet mean \citep{willis2019confidence}, which could provide better visualizations of uncertainty and more information about plausible underlying tree structures.

Finally, substantial work remains in prompt design, output embeddings, and benchmark construction. In particular, we need prompts that expose clear and meaningful characteristics of LLM behavior so that these analyses can be aligned more closely with human-created benchmarks.

\section*{Acknowledgments}
We used LLMs for minor polishing of the text and for coding assistance, such as writing scripts to run batch experiments.

\section*{Copyright}
Copyright 2026 Carnegie Mellon University.

This material is based upon work supported by the Department of War under Air Force Contract No. FA8702-15-D-0002 with Carnegie Mellon University for the operation of the Software Engineering Institute, a federally funded research and development center.  

The opinions, findings, conclusions, and/or recommendations contained in this material are those of the author(s) and should not be construed as an official US Government position, policy, or decision, unless designated by other documentation.  

References herein to any specific entity, product, process, or service by trade name, trademark, manufacturer, or otherwise, does not necessarily constitute or imply its endorsement, recommendation, or favoring by Carnegie Mellon University or its Software Engineering Institute nor of Carnegie Mellon University - Software Engineering Institute by any such named or represented entity.

NO WARRANTY. THIS CARNEGIE MELLON UNIVERSITY AND SOFTWARE ENGINEERING INSTITUTE MATERIAL IS FURNISHED ON AN "AS-IS" BASIS. CARNEGIE MELLON UNIVERSITY MAKES NO WARRANTIES OF ANY KIND, EITHER EXPRESSED OR IMPLIED, AS TO ANY MATTER INCLUDING, BUT NOT LIMITED TO, WARRANTY OF FITNESS FOR PURPOSE OR MERCHANTABILITY, EXCLUSIVITY, OR RESULTS OBTAINED FROM USE OF THE MATERIAL. CARNEGIE MELLON UNIVERSITY DOES NOT MAKE ANY WARRANTY OF ANY KIND WITH RESPECT TO FREEDOM FROM PATENT, TRADEMARK, OR COPYRIGHT INFRINGEMENT.

[DISTRIBUTION STATEMENT A] This material has been approved for public release and unlimited distribution.  Please see Copyright notice for non-US Government use and distribution.

This work is licensed under a Creative Commons Attribution-NonCommercial 4.0 International License (https://creativecommons.org/licenses/by-nc/4.0/).  Requests for permission for non-licensed uses should be directed to the Software Engineering Institute at permission@sei.cmu.edu.

DM26-0361

\bibliographystyle{plainnat}
\bibliography{references}

\appendix

\section{Model, Data, and Prompts}\label{app:data}

\subsection{Controlled Experiment}

The primary model used in our controlled experiment is T5-small, available on Hugging Face at \url{https://huggingface.co/google-t5/t5-small} \citep{2020t5}.

The datasets used to fine-tune T5-small are listed in Table \ref{tab:datasets}. For each dataset, we use the first 10,000 rows and split them into training and test sets for fine-tuning.

\setlength{\tabcolsep}{4pt}
\renewcommand{\arraystretch}{1.12}

\begin{table}[ht]
\footnotesize
\centering
\caption{Data used to fine-tune T5-small. All datasets can be accessed at \url{https://huggingface.co/datasets/<dataset_name>}.}
\label{tab:datasets}
\begin{threeparttable}
\begin{tabularx}{\textwidth}{@{} l >{\raggedright\arraybackslash}X r @{}}
\toprule
\textbf{Dataset} & \textbf{Description (citation)} & \textbf{Rows (total)} \\
\midrule
BillSum (BS)        & ``Summarization of US Congressional and California state bills.'' \citep{Kornilova2019BillSum}. & 23{,}455 \\
XSum (XS)           & ``Extreme'' single-sentence summaries of BBC news articles \citep{Narayan2018XSum}. & 226{,}711 \\
CNN/DailyMail v1.0.0 (CD)  & News article to short highlights \citep{Hermann2015CNNDM,see-etal-2017-get}. & 311{,}971 \\
arXiv Summ. (AX)    & Long scientific paper summaries (article $\rightarrow$ abstract) \citep{Cohan2018LongSumm}. & 215{,}913 \\
BIGPATENT (BP)      & Large-scale patent description to abstract summarization \citep{Sharma2019BigPatent}. & 1{,}341{,}362 \\
DialogSum (DS)      & Real-life dialogue transcripts paired with concise summaries \citep{Chen2021DialogSum}. & 14{,}460 \\
GovReport (GR)      & Long U.S. government reports with human summaries \citep{Huang2021GovReport}. & 19{,}463 \\
PubMed Summ. (PM)   & Biomedical paper summaries (article $\rightarrow$ abstract) \citep{Cohan2018LongSumm}. & 133{,}215 \\
SAMSum (SS)         & Chat/IM-style conversations with human-written summaries \citep{Gliwa2019SAMSum}. & 16{,}369 \\
DebateSum (DeS)     & Competitive debate evidence with very short abstracts \citep{RoushBalaji2020DebateSum}. & 240{,}566 \\
\bottomrule
\end{tabularx}
\begin{tablenotes}[flushleft]
\footnotesize\ttfamily
\item[] HF paths (post \texttt{datasets/}): BS=FiscalNote/billsum; XS=EdinburghNLP/xsum; CD=abisee/cnn\_dailymail; AX=ccdv/arxiv-summarization; BP=NortheasternUniversity/big\_patent; DS=knkarthick/dialogsum; GR=ccdv/govreport-summarization; PM=ccdv/pubmed-summarization; SS=knkarthick/samsum; DeS=Hellisotherpeople/DebateSum.
\end{tablenotes}
\end{threeparttable}
\end{table}

Our input-target pair for fine-tuning is \texttt{summarize: <doc>} as input and \texttt{<summary>} as target, where the full document and its summary are inserted as appropriate. The maximum input length is 1024 tokens and the maximum target length is 128 tokens, so the summaries are kept short.

\subsection{Unsupervised Experiment}

In Section \ref{sec:output-embeddings}, we present an evolutionary tree estimated from output-text embeddings produced by different foundation models. The models used in this experiment are shown in Figure \ref{fig:pca-and-tree-embeddings}. For each model, we prepend the phrase ``Summarize the following text in $<$100 words:'' to the document. Responses are limited to a maximum of 256 tokens. Using the OpenAI API for the four OpenAI models shown in this figure, the experiments cost \$5.86.

\section{Additional Experimental Details and Graphs}\label{app:pca-all}

\begin{figure}
    \centering
    \includegraphics[width=1\linewidth]{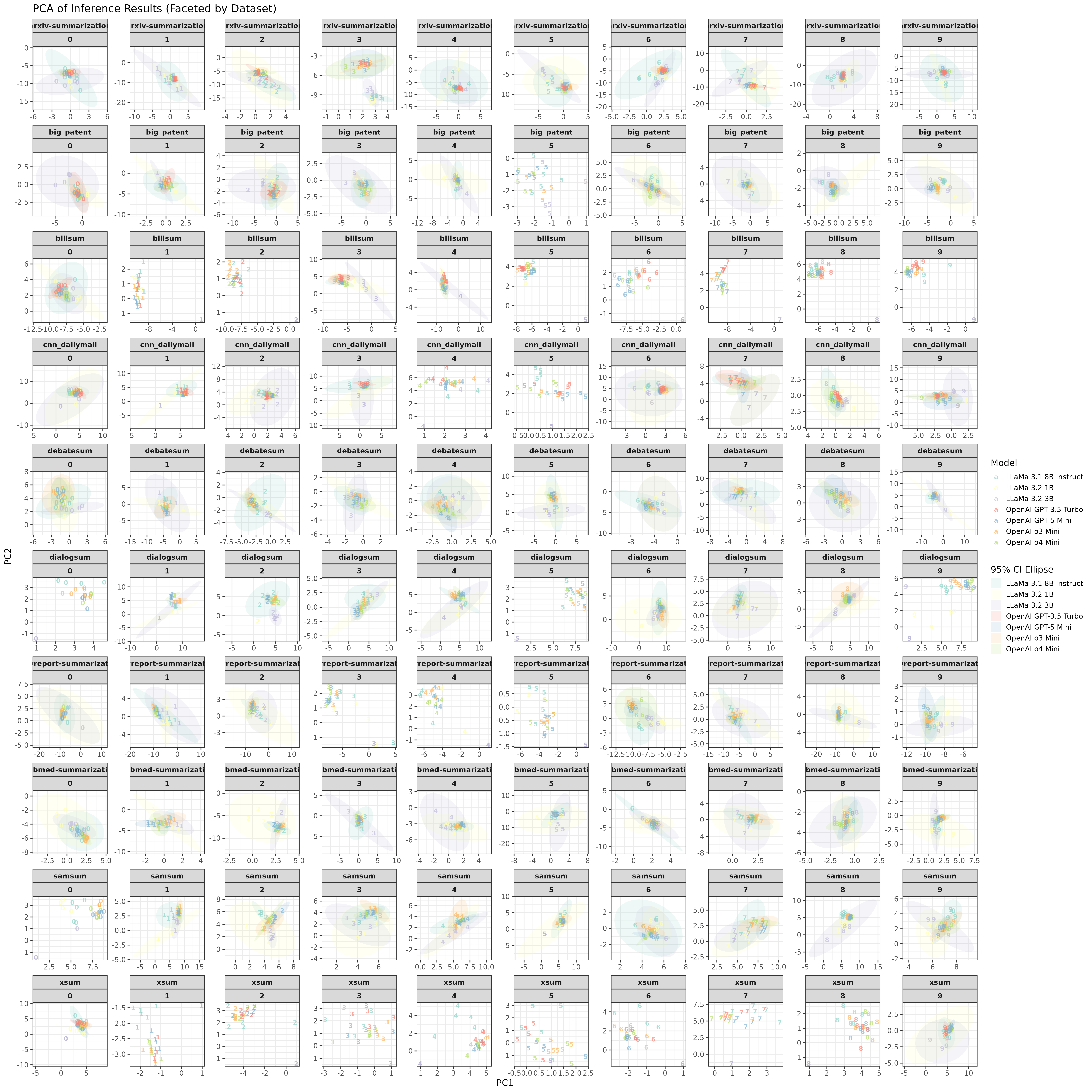}
    \caption{PCA shown by dataset and prompt for each of the foundation models.}
    \label{fig:llama-openai-pca-dataset-prompt}
\end{figure}

We show PCA visualizations for each dataset-prompt pair for the foundation models in Figure \ref{fig:llama-openai-pca-dataset-prompt}. In general, some separation among the models is visible, although it is not always pronounced. Note that the scales differ across panels in order to better show clustering.

\section{Robustness of Phylogenies Across All Nodes}\label{app:robust}

Here we report RF scores for evolutionary trees estimated from all pairwise model distances, rather than only the leaf nodes. Because phylogenetic estimation assigns all observed objects to leaves, RF scores are naturally larger in this setting than in the leaf-only analysis.

Table \ref{tab:tree_stats-all-nodes} is analogous to Table \ref{tab:tree_stats-all}, but it reports results for trees estimated using all nodes rather than only leaves.

For single-tree estimates, RF scores are substantially larger than in the leaf-only analysis. For consensus trees estimated from layer-wise distances, RF distances are much smaller and only moderately larger than in the leaf-only case. Note also that Match (\%) values are much smaller here, suggesting substantially greater variability in the estimated trees. Even so, Random RF $<$ Consensus RF remains below 0.05, indicating that the consensus trees are still reliable estimates of the underlying topology.

\begin{table}[!ht]
\centering
\caption{\label{tab:tree_stats-all-nodes}Average values over the 50 experiments.}
\resizebox{\ifdim\width>\linewidth\linewidth\else\width\fi}{!}{
\begin{tabular}[t]{rrrlrrrr}
\toprule
\# Leaves & \# Nodes & n & Metric & Total Weight RF (SD) & Consensus Weight RF (SD) & Match (\%) & Random RF $<$ Consensus RF\\
\midrule
\cellcolor{gray!10}{3} & \cellcolor{gray!10}{6} & \cellcolor{gray!10}{4} & \cellcolor{gray!10}{Correlation} & \cellcolor{gray!10}{4.25 (1.71)} & \cellcolor{gray!10}{0.00 (0.00)} & \cellcolor{gray!10}{46.0} & \cellcolor{gray!10}{0.000}\\
3 & 6 & 4 & Cosine & 4.50 (1.73) & 0.00 (0.00) & 41.6 & 0.000\\
\cellcolor{gray!10}{3} & \cellcolor{gray!10}{6} & \cellcolor{gray!10}{4} & \cellcolor{gray!10}{Threshold} & \cellcolor{gray!10}{5.25 (0.96)} & \cellcolor{gray!10}{0.00 (0.00)} & \cellcolor{gray!10}{28.1} & \cellcolor{gray!10}{0.000}\\
3 & 6 & 4 & \(L_1\) & 5.00 (1.15) & 0.00 (0.00) & 60.9 & 0.000\\
\cellcolor{gray!10}{3} & \cellcolor{gray!10}{6} & \cellcolor{gray!10}{4} & \cellcolor{gray!10}{\(L_2\)} & \cellcolor{gray!10}{5.00 (1.15)} & \cellcolor{gray!10}{0.00 (0.00)} & \cellcolor{gray!10}{60.5} & \cellcolor{gray!10}{0.000}\\
\addlinespace
4 & 8 & 1 & Correlation & NA & NA & 31.3 & 0.000\\
\cellcolor{gray!10}{4} & \cellcolor{gray!10}{8} & \cellcolor{gray!10}{1} & \cellcolor{gray!10}{Cosine} & \cellcolor{gray!10}{NA} & \cellcolor{gray!10}{NA} & \cellcolor{gray!10}{32.1} & \cellcolor{gray!10}{0.000}\\
4 & 8 & 1 & Threshold & NA & NA & 16.0 & 0.000\\
\cellcolor{gray!10}{4} & \cellcolor{gray!10}{8} & \cellcolor{gray!10}{1} & \cellcolor{gray!10}{\(L_1\)} & \cellcolor{gray!10}{NA} & \cellcolor{gray!10}{NA} & \cellcolor{gray!10}{19.1} & \cellcolor{gray!10}{0.000}\\
4 & 8 & 1 & \(L_2\) & NA & NA & 21.4 & 0.000\\
\addlinespace
\cellcolor{gray!10}{4} & \cellcolor{gray!10}{9} & \cellcolor{gray!10}{1} & \cellcolor{gray!10}{Correlation} & \cellcolor{gray!10}{10.00 (1.41)} & \cellcolor{gray!10}{NA} & \cellcolor{gray!10}{38.9} & \cellcolor{gray!10}{0.000}\\
4 & 9 & 1 & Cosine & 11.00 (1.41) & NA & 39.7 & 0.000\\
\cellcolor{gray!10}{4} & \cellcolor{gray!10}{9} & \cellcolor{gray!10}{1} & \cellcolor{gray!10}{Threshold} & \cellcolor{gray!10}{9.00 (0.00)} & \cellcolor{gray!10}{NA} & \cellcolor{gray!10}{42.7} & \cellcolor{gray!10}{0.000}\\
4 & 9 & 1 & \(L_1\) & 8.50 (0.71) & NA & 30.5 & 0.000\\
\cellcolor{gray!10}{4} & \cellcolor{gray!10}{9} & \cellcolor{gray!10}{1} & \cellcolor{gray!10}{\(L_2\)} & \cellcolor{gray!10}{7.50 (2.12)} & \cellcolor{gray!10}{NA} & \cellcolor{gray!10}{29.0} & \cellcolor{gray!10}{0.000}\\
\addlinespace
5 & 9 & 1 & Correlation & 10.00 (1.41) & NA & 29.0 & 0.000\\
\cellcolor{gray!10}{5} & \cellcolor{gray!10}{9} & \cellcolor{gray!10}{1} & \cellcolor{gray!10}{Cosine} & \cellcolor{gray!10}{11.00 (1.41)} & \cellcolor{gray!10}{NA} & \cellcolor{gray!10}{26.7} & \cellcolor{gray!10}{0.000}\\
5 & 9 & 1 & Threshold & 9.00 (0.00) & NA & 11.5 & 0.000\\
\cellcolor{gray!10}{5} & \cellcolor{gray!10}{9} & \cellcolor{gray!10}{1} & \cellcolor{gray!10}{\(L_1\)} & \cellcolor{gray!10}{8.50 (0.71)} & \cellcolor{gray!10}{NA} & \cellcolor{gray!10}{45.8} & \cellcolor{gray!10}{0.000}\\
5 & 9 & 1 & \(L_2\) & 7.50 (2.12) & NA & 42.0 & 0.000\\
\addlinespace
\cellcolor{gray!10}{5} & \cellcolor{gray!10}{10} & \cellcolor{gray!10}{4} & \cellcolor{gray!10}{Correlation} & \cellcolor{gray!10}{13.25 (1.71)} & \cellcolor{gray!10}{0.50 (1.00)} & \cellcolor{gray!10}{15.8} & \cellcolor{gray!10}{0.067}\\
5 & 10 & 4 & Cosine & 13.75 (1.26) & 0.50 (1.00) & 13.7 & 0.067\\
\cellcolor{gray!10}{5} & \cellcolor{gray!10}{10} & \cellcolor{gray!10}{4} & \cellcolor{gray!10}{Threshold} & \cellcolor{gray!10}{6.00 (5.83)} & \cellcolor{gray!10}{0.50 (1.00)} & \cellcolor{gray!10}{9.5} & \cellcolor{gray!10}{0.067}\\
5 & 10 & 4 & \(L_1\) & 10.75 (1.50) & 0.50 (1.00) & 24.2 & 0.067\\
\cellcolor{gray!10}{5} & \cellcolor{gray!10}{10} & \cellcolor{gray!10}{4} & \cellcolor{gray!10}{\(L_2\)} & \cellcolor{gray!10}{11.00 (1.15)} & \cellcolor{gray!10}{0.50 (1.00)} & \cellcolor{gray!10}{22.5} & \cellcolor{gray!10}{0.067}\\
\addlinespace
5 & 11 & 11 & Correlation & 14.76 (1.36) & 0.18 (0.60) & 19.8 & 0.029\\
\cellcolor{gray!10}{5} & \cellcolor{gray!10}{11} & \cellcolor{gray!10}{11} & \cellcolor{gray!10}{Cosine} & \cellcolor{gray!10}{14.65 (1.38)} & \cellcolor{gray!10}{0.09 (0.30)} & \cellcolor{gray!10}{18.6} & \cellcolor{gray!10}{0.029}\\
5 & 11 & 11 & Threshold & 10.92 (2.72) & 0.09 (0.30) & 9.0 & 0.029\\
\cellcolor{gray!10}{5} & \cellcolor{gray!10}{11} & \cellcolor{gray!10}{11} & \cellcolor{gray!10}{\(L_1\)} & \cellcolor{gray!10}{12.24 (1.44)} & \cellcolor{gray!10}{0.09 (0.30)} & \cellcolor{gray!10}{27.3} & \cellcolor{gray!10}{0.029}\\
5 & 11 & 11 & \(L_2\) & 12.11 (1.78) & 0.09 (0.30) & 26.2 & 0.029\\
\addlinespace
\cellcolor{gray!10}{6} & \cellcolor{gray!10}{11} & \cellcolor{gray!10}{26} & \cellcolor{gray!10}{Correlation} & \cellcolor{gray!10}{14.76 (1.36)} & \cellcolor{gray!10}{0.42 (0.81)} & \cellcolor{gray!10}{14.2} & \cellcolor{gray!10}{0.037}\\
6 & 11 & 26 & Cosine & 14.65 (1.38) & 0.42 (0.81) & 13.9 & 0.037\\
\cellcolor{gray!10}{6} & \cellcolor{gray!10}{11} & \cellcolor{gray!10}{26} & \cellcolor{gray!10}{Threshold} & \cellcolor{gray!10}{10.92 (2.72)} & \cellcolor{gray!10}{0.38 (0.75)} & \cellcolor{gray!10}{9.2} & \cellcolor{gray!10}{0.037}\\
6 & 11 & 26 & \(L_1\) & 12.24 (1.44) & 0.38 (0.70) & 20.0 & 0.043\\
\cellcolor{gray!10}{6} & \cellcolor{gray!10}{11} & \cellcolor{gray!10}{26} & \cellcolor{gray!10}{\(L_2\)} & \cellcolor{gray!10}{12.11 (1.78)} & \cellcolor{gray!10}{0.35 (0.69)} & \cellcolor{gray!10}{19.1} & \cellcolor{gray!10}{0.037}\\
\bottomrule
\end{tabular}}
\end{table}

\end{document}

%% file: math_commands.tex
\usepackage{amsmath,amsfonts,bm}

\def\eqref#1{equation~\ref{#1}}

\def\1{\bm{1}}

\DeclareMathAlphabet{\mathsfit}{\encodingdefault}{\sfdefault}{m}{sl}
\SetMathAlphabet{\mathsfit}{bold}{\encodingdefault}{\sfdefault}{bx}{n}



%% file: references.bib
@article{yang2023harnessing,
    title={Harnessing the Power of LLMs in Practice: A Survey on ChatGPT and Beyond}, 
    author={Jingfeng Yang and Hongye Jin and Ruixiang Tang and Xiaotian Han and Qizhang Feng and Haoming Jiang and Bing Yin and Xia Hu},
    year={2023},
    eprint={2304.13712},
    archivePrefix={arXiv},
    primaryClass={cs.CL}
}

@article{vaswani2017attention,
  title={Attention is all you need},
  author={Vaswani, Ashish and Shazeer, Noam and Parmar, Niki and Uszkoreit, Jakob and Jones, Llion and Gomez, Aidan N and Kaiser, {\L}ukasz and Polosukhin, Illia},
  journal={Advances in neural information processing systems},
  volume={30},
  year={2017}
}

@article{
liang2023holistic,
title={Holistic Evaluation of Language Models},
author={Percy Liang and Rishi Bommasani and Tony Lee and Dimitris Tsipras and Dilara Soylu and Michihiro Yasunaga and Yian Zhang and Deepak Narayanan and Yuhuai Wu and Ananya Kumar and Benjamin Newman and Binhang Yuan and Bobby Yan and Ce Zhang and Christian Alexander Cosgrove and Christopher D Manning and Christopher Re and Diana Acosta-Navas and Drew Arad Hudson and Eric Zelikman and Esin Durmus and Faisal Ladhak and Frieda Rong and Hongyu Ren and Huaxiu Yao and Jue WANG and Keshav Santhanam and Laurel Orr and Lucia Zheng and Mert Yuksekgonul and Mirac Suzgun and Nathan Kim and Neel Guha and Niladri S. Chatterji and Omar Khattab and Peter Henderson and Qian Huang and Ryan Andrew Chi and Sang Michael Xie and Shibani Santurkar and Surya Ganguli and Tatsunori Hashimoto and Thomas Icard and Tianyi Zhang and Vishrav Chaudhary and William Wang and Xuechen Li and Yifan Mai and Yuhui Zhang and Yuta Koreeda},
journal={Transactions on Machine Learning Research},
issn={2835-8856},
year={2023},
url={https://openreview.net/forum?id=iO4LZibEqW},
note={Featured Certification, Expert Certification}
}

@misc{hendrycks2021measuringmassivemultitasklanguage,
      title={Measuring Massive Multitask Language Understanding}, 
      author={Dan Hendrycks and Collin Burns and Steven Basart and Andy Zou and Mantas Mazeika and Dawn Song and Jacob Steinhardt},
      year={2021},
      eprint={2009.03300},
      archivePrefix={arXiv},
      primaryClass={cs.CY},
      url={https://arxiv.org/abs/2009.03300}, 
}

@misc{wang2019gluemultitaskbenchmarkanalysis,
      title={GLUE: A Multi-Task Benchmark and Analysis Platform for Natural Language Understanding}, 
      author={Alex Wang and Amanpreet Singh and Julian Michael and Felix Hill and Omer Levy and Samuel R. Bowman},
      year={2019},
      eprint={1804.07461},
      archivePrefix={arXiv},
      primaryClass={cs.CL},
      url={https://arxiv.org/abs/1804.07461}, 
}

@inproceedings{lin2004rouge,
  title={Rouge: A package for automatic evaluation of summaries},
  author={Lin, Chin-Yew},
  booktitle={Text summarization branches out},
  pages={74--81},
  year={2004}
}

@inproceedings{banerjee2005meteor,
  title={METEOR: An automatic metric for MT evaluation with improved correlation with human judgments},
  author={Banerjee, Satanjeev and Lavie, Alon},
  booktitle={Proceedings of the acl workshop on intrinsic and extrinsic evaluation measures for machine translation and/or summarization},
  pages={65--72},
  year={2005}
}

@inproceedings{papineni2002bleu,
  title={Bleu: a method for automatic evaluation of machine translation},
  author={Papineni, Kishore and Roukos, Salim and Ward, Todd and Zhu, Wei-Jing},
  booktitle={Proceedings of the 40th annual meeting of the Association for Computational Linguistics},
  pages={311--318},
  year={2002}
}

@misc{chiang2024chatbotarenaopenplatform,
      title={Chatbot Arena: An Open Platform for Evaluating LLMs by Human Preference}, 
      author={Wei-Lin Chiang and Lianmin Zheng and Ying Sheng and Anastasios Nikolas Angelopoulos and Tianle Li and Dacheng Li and Hao Zhang and Banghua Zhu and Michael Jordan and Joseph E. Gonzalez and Ion Stoica},
      year={2024},
      eprint={2403.04132},
      archivePrefix={arXiv},
      primaryClass={cs.AI},
      url={https://arxiv.org/abs/2403.04132}, 
}

@misc{li2024llmsasjudgescomprehensivesurveyllmbased,
      title={LLMs-as-Judges: A Comprehensive Survey on LLM-based Evaluation Methods}, 
      author={Haitao Li and Qian Dong and Junjie Chen and Huixue Su and Yujia Zhou and Qingyao Ai and Ziyi Ye and Yiqun Liu},
      year={2024},
      eprint={2412.05579},
      archivePrefix={arXiv},
      primaryClass={cs.CL},
      url={https://arxiv.org/abs/2412.05579}, 
}

@inproceedings{gallagher2024assessing,
  title={Assessing llms for high stakes applications},
  author={Gallagher, Shannon K and Ratchford, Jasmine and Brooks, Tyler and Brown, Bryan P and Heim, Eric and Nichols, William R and Mcmillan, Scott and Rallapalli, Swati and Smith, Carol J and VanHoudnos, Nathan and others},
  booktitle={Proceedings of the 46th International Conference on Software Engineering: Software Engineering in Practice},
  pages={103--105},
  year={2024}
}

@misc{wang2024mmluprorobustchallengingmultitask,
      title={MMLU-Pro: A More Robust and Challenging Multi-Task Language Understanding Benchmark}, 
      author={Yubo Wang and Xueguang Ma and Ge Zhang and Yuansheng Ni and Abhranil Chandra and Shiguang Guo and Weiming Ren and Aaran Arulraj and Xuan He and Ziyan Jiang and Tianle Li and Max Ku and Kai Wang and Alex Zhuang and Rongqi Fan and Xiang Yue and Wenhu Chen},
      year={2024},
      eprint={2406.01574},
      archivePrefix={arXiv},
      primaryClass={cs.CL},
      url={https://arxiv.org/abs/2406.01574}, 
}

@article{kiela2021dynabench,
  title={Dynabench: Rethinking benchmarking in NLP},
  author={Kiela, Douwe and Bartolo, Max and Nie, Yixin and Kaushik, Divyansh and Geiger, Atticus and Wu, Zhengxuan and Vidgen, Bertie and Prasad, Grusha and Singh, Amanpreet and Ringshia, Pratik and others},
  journal={arXiv preprint arXiv:2104.14337},
  year={2021}
}

@article{zhang2019bertscore,
  title={Bertscore: Evaluating text generation with bert},
  author={Zhang, Tianyi and Kishore, Varsha and Wu, Felix and Weinberger, Kilian Q and Artzi, Yoav},
  journal={arXiv preprint arXiv:1904.09675},
  year={2019}
}

@article{laban2022summac,
  title={SummaC: Re-visiting NLI-based models for inconsistency detection in summarization},
  author={Laban, Philippe and Schnabel, Tobias and Bennett, Paul N and Hearst, Marti A},
  journal={Transactions of the Association for Computational Linguistics},
  volume={10},
  pages={163--177},
  year={2022},
  publisher={MIT Press One Rogers Street, Cambridge, MA 02142-1209, USA journals-info~…}
}

@article{grusky2018newsroom,
  title={Newsroom: A dataset of 1.3 million summaries with diverse extractive strategies},
  author={Grusky, Max and Naaman, Mor and Artzi, Yoav},
  journal={arXiv preprint arXiv:1804.11283},
  year={2018}
}

@article{cambria2024xai,
  title={Xai meets llms: A survey of the relation between explainable ai and large language models},
  author={Cambria, Erik and Malandri, Lorenzo and Mercorio, Fabio and Nobani, Navid and Seveso, Andrea},
  journal={arXiv preprint arXiv:2407.15248},
  year={2024}
}

@inproceedings{macukow2016neural,
  title={Neural networks--state of art, brief history, basic models and architecture},
  author={Macukow, Bohdan},
  booktitle={IFIP international conference on computer information systems and industrial management},
  pages={3--14},
  year={2016},
  organization={Springer}
}

@misc{lindsey_et_al_2025_biology,
  title  = {On the Biology of a Large Language Model},
  author = {Lindsey, Jack and Gurnee, Wes and Ameisen, Emmanuel and Chen, Brian and Pearce, Adam and Turner, Nicholas L. and Citro, Craig and Abrahams, David and Carter, Shan and Hosmer, Basil and Marcus, Jonathan and Sklar, Michael and Templeton, Adly and Bricken, Trenton and McDougall, Callum and Cunningham, Hoagy and Henighan, Thomas and Jermyn, Adam and Jones, Andy and Persic, Andrew and Qi, Zhenyi and Thompson, T. Ben and Zimmerman, Sam and Rivoire, Kelley and Conerly, Thomas and Olah, Chris and Batson, Joshua},
  year   = {2025},
  month  = mar,
  note   = {Accessed on \textit{August 19, 2025}.},  
  howpublished = {\url{https://transformer-circuits.pub/2025/attribution-graphs/biology.html}},
}

@misc{wu2025usablexai10strategies,
      title={Usable XAI: 10 Strategies Towards Exploiting Explainability in the LLM Era}, 
      author={Xuansheng Wu and Haiyan Zhao and Yaochen Zhu and Yucheng Shi and Fan Yang and Lijie Hu and Tianming Liu and Xiaoming Zhai and Wenlin Yao and Jundong Li and Mengnan Du and Ninghao Liu},
      year={2025},
      eprint={2403.08946},
      archivePrefix={arXiv},
      primaryClass={cs.LG},
      url={https://arxiv.org/abs/2403.08946}, 
}

@article{cloud2025subliminal,
  title={Subliminal Learning: Language models transmit behavioral traits via hidden signals in data},
  author={Cloud, Alex and Le, Minh and Chua, James and Betley, Jan and Sztyber-Betley, Anna and Hilton, Jacob and Marks, Samuel and Evans, Owain},
  journal={arXiv preprint arXiv:2507.14805},
  year={2025}
}

@article{turista2020distribution,
  title={Distribution of COVID-19 and phylogenetic tree construction of SARS-CoV-2 in Indonesia},
  author={Turista, Dora Dayu Rahma and Islamy, Aesthetica and Kharisma, Viol Dhea and Ansori, Arif Nur Muhammad},
  journal={J Pure Appl Microbiol},
  volume={14},
  number={suppl 1},
  pages={1035--42},
  year={2020}
}

@article{birky1995uniparental,
  title={Uniparental inheritance of mitochondrial and chloroplast genes: mechanisms and evolution.},
  author={Birky Jr, C William},
  journal={Proceedings of the National Academy of Sciences},
  volume={92},
  number={25},
  pages={11331--11338},
  year={1995}
}

@article{graham2018phylogenetic,
  title={Phylogenetic scale in ecology and evolution},
  author={Graham, Catherine H and Storch, David and Machac, Antonin},
  journal={Global Ecology and Biogeography},
  volume={27},
  number={2},
  pages={175--187},
  year={2018},
  publisher={Wiley Online Library}
}

@article{morrison2014phylogenetic,
  title={Phylogenetic networks: a review of methods to display evolutionary history},
  author={Morrison, David A},
  journal={Annual Research \& Review in Biology},
  volume={4},
  number={10},
  pages={1518},
  year={2014},
  publisher={SCIENCEDOMAIN International}
}

@article{mirjalili2019evolutionary,
  title={Evolutionary algorithms and neural networks},
  author={Mirjalili, Seyedali},
  journal={Studies in computational intelligence},
  volume={780},
  number={1},
  pages={43--53},
  year={2019},
  publisher={Springer}
}

@inproceedings{Kornilova2019BillSum,
    title = "{B}ill{S}um: A Corpus for Automatic Summarization of {US} Legislation",
    author = "Kornilova, Anastassia  and
      Eidelman, Vladimir",
    editor = "Wang, Lu  and
      Cheung, Jackie Chi Kit  and
      Carenini, Giuseppe  and
      Liu, Fei",
    booktitle = "Proceedings of the 2nd Workshop on New Frontiers in Summarization",
    month = nov,
    year = "2019",
    address = "Hong Kong, China",
    publisher = "Association for Computational Linguistics",
    url = "https://aclanthology.org/D19-5406",
    doi = "10.18653/v1/D19-5406",
    pages = "48--56",
    eprint={1910.00523},
    archivePrefix={arXiv},
    primaryClass={cs.CL},
}

@article{Narayan2018XSum,
  title={Don't Give Me the Details, Just the Summary! Topic-Aware Convolutional Neural Networks for Extreme Summarization},
  author={Shashi Narayan and Shay B. Cohen and Mirella Lapata},
  journal={ArXiv},
  year={2018},
  volume={abs/1808.08745}
}

@inproceedings{Hermann2015CNNDM,
  title={Teaching machines to read and comprehend},
  author={Hermann, Karl Moritz and Kocisky, Tomas and Grefenstette, Edward and Espeholt, Lasse and Kay, Will and Suleyman, Mustafa and Blunsom, Phil},
  journal={Advances in neural information processing systems},
  volume={28},
  year={2015}
}

@inproceedings{see-etal-2017-get,
    title = "Get To The Point: Summarization with Pointer-Generator Networks",
    author = "See, Abigail  and
      Liu, Peter J.  and
      Manning, Christopher D.",
    booktitle = "Proceedings of the 55th Annual Meeting of the Association for Computational Linguistics (Volume 1: Long Papers)",
    month = jul,
    year = "2017",
    address = "Vancouver, Canada",
    publisher = "Association for Computational Linguistics",
    url = "https://www.aclweb.org/anthology/P17-1099",
    doi = "10.18653/v1/P17-1099",
    pages = "1073--1083",
}

@inproceedings{Cohan2018LongSumm,
  title = "A Discourse-Aware Attention Model for Abstractive Summarization of Long Documents",
  author = "Cohan, Arman  and
    Dernoncourt, Franck  and
    Kim, Doo Soon  and
    Bui, Trung  and
    Kim, Seokhwan  and
    Chang, Walter  and
    Goharian, Nazli",
  booktitle = "Proceedings of the 2018 Conference of the North {A}merican Chapter of the Association for Computational Linguistics: Human Language Technologies, Volume 2 (Short Papers)",
  month = jun,
  year = "2018",
  address = "New Orleans, Louisiana",
  publisher = "Association for Computational Linguistics",
  url = "https://aclanthology.org/N18-2097",
  doi = "10.18653/v1/N18-2097",
  pages = "615--621",
}

@article{Sharma2019BigPatent,
  author    = {Eva Sharma and
               Chen Li and
               Lu Wang},
  title     = {{BIGPATENT:} {A} Large-Scale Dataset for Abstractive and Coherent
               Summarization},
  journal   = {CoRR},
  volume    = {abs/1906.03741},
  year      = {2019},
  url       = {http://arxiv.org/abs/1906.03741},
  eprinttype = {arXiv},
  eprint    = {1906.03741},
  bibsource = {dblp computer science bibliography, https://dblp.org}
}

@inproceedings{Chen2021DialogSum,
    title = "{D}ialog{S}um: {A} Real-Life Scenario Dialogue Summarization Dataset",
    author = "Chen, Yulong  and
      Liu, Yang  and
      Chen, Liang  and
      Zhang, Yue",
    booktitle = "Findings of the Association for Computational Linguistics: ACL-IJCNLP 2021",
    month = aug,
    year = "2021",
    address = "Online",
    publisher = "Association for Computational Linguistics",
    url = "https://aclanthology.org/2021.findings-acl.449",
    doi = "10.18653/v1/2021.findings-acl.449",
    pages = "5062--5074"
}

@misc{Huang2021GovReport,
      title={Efficient Attentions for Long Document Summarization}, 
      author={Luyang Huang and Shuyang Cao and Nikolaus Parulian and Heng Ji and Lu Wang},
      year={2021},
      eprint={2104.02112},
      archivePrefix={arXiv},
      primaryClass={cs.CL}
}

@inproceedings{Gliwa2019SAMSum,
    title = "{SAMS}um Corpus: A Human-annotated Dialogue Dataset for Abstractive Summarization",
    author = "Gliwa, Bogdan  and
      Mochol, Iwona  and
      Biesek, Maciej  and
      Wawer, Aleksander",
    booktitle = "Proceedings of the 2nd Workshop on New Frontiers in Summarization",
    month = nov,
    year = "2019",
    address = "Hong Kong, China",
    publisher = "Association for Computational Linguistics",
    url = "https://www.aclweb.org/anthology/D19-5409",
    doi = "10.18653/v1/D19-5409",
    pages = "70--79"
}

@inproceedings{RoushBalaji2020DebateSum,
    title = "{D}ebate{S}um: A large-scale argument mining and summarization dataset",
    author = "Roush, Allen  and
      Balaji, Arvind",
    editor = "Cabrio, Elena  and
      Villata, Serena",
    booktitle = "Proceedings of the 7th Workshop on Argument Mining",
    month = dec,
    year = "2020",
    address = "Online",
    publisher = "Association for Computational Linguistics",
    url = "https://aclanthology.org/2020.argmining-1.1/",
    pages = "1--7"
}

@article{2020t5,
  author  = {Colin Raffel and Noam Shazeer and Adam Roberts and Katherine Lee and Sharan Narang and Michael Matena and Yanqi Zhou and Wei Li and Peter J. Liu},
  title   = {Exploring the Limits of Transfer Learning with a Unified Text-to-Text Transformer},
  journal = {Journal of Machine Learning Research},
  year    = {2020},
  volume  = {21},
  number  = {140},
  pages   = {1-67},
  url     = {http://jmlr.org/papers/v21/20-074.html}
}

@article{njsaitou1987,
    author = {Saitou, N and Nei, M},
    title = {The neighbor-joining method: a new method for reconstructing phylogenetic trees.},
    journal = {Molecular Biology and Evolution},
    volume = {4},
    number = {4},
    pages = {406-425},
    year = {1987},
    month = {07},
    issn = {0737-4038},
    doi = {10.1093/oxfordjournals.molbev.a040454},
    url = {https://doi.org/10.1093/oxfordjournals.molbev.a040454},
    eprint = {https://academic.oup.com/mbe/article-pdf/4/4/406/11167444/7sait.pdf},
}

@article{robinson1981comparison,
  title={Comparison of phylogenetic trees},
  author={Robinson, David F and Foulds, Leslie R},
  journal={Mathematical biosciences},
  volume={53},
  number={1-2},
  pages={131--147},
  year={1981},
  publisher={Elsevier}
}

@inproceedings{lin2003automatic,
  title={Automatic evaluation of summaries using n-gram co-occurrence statistics},
  author={Lin, Chin-Yew and Hovy, Eduard},
  booktitle={Proceedings of the 2003 human language technology conference of the North American chapter of the association for computational linguistics},
  pages={150--157},
  year={2003}
}

@article{willis2019confidence,
  title={Confidence sets for phylogenetic trees},
  author={Willis, Amy},
  journal={Journal of the American Statistical Association},
  volume={114},
  number={525},
  pages={235--244},
  year={2019},
  publisher={Taylor \& Francis}
}

@article{yax2024phylolm,
  title={PhyloLM: Inferring the phylogeny of large language models and predicting their performances in benchmarks},
  author={Yax, Nicolas and Oudeyer, Pierre-Yves and Palminteri, Stefano},
  journal={arXiv preprint arXiv:2404.04671},
  year={2024}
}

@article{horwitz2024unsupervised,
  title={Unsupervised model tree heritage recovery},
  author={Horwitz, Eliahu and Shul, Asaf and Hoshen, Yedid},
  journal={arXiv preprint arXiv:2405.18432},
  year={2024}
}
